# Rethinking Deconvolution for 2D Human Pose Estimation Light yet Accurate Model for Real-time Edge Computing

Masayuki Yamazaki[1] and Eigo Mori[2]
[1] Toyota Motor Corporation, Japan, masayuki_yamazaki_aa@mail.toyota.co.jp
[2] Sigfoss Corporation, Japan, eigo@sigfoss.com

*Abstract*—In this study, we present a pragmatic lightweight pose estimation model. Our model can achieve real-time predictions using low-power embedded devices. This system was found to be very accurate and achieved a 94.5% accuracy of SOTA HRNet 256×192 using a computational cost of only 3.8% on COCO test dataset. Our model adopts an encoder-decoder architecture and is carefully downsized to improve its efficiency. We especially focused on optimizing the deconvolution layers and observed that the channel reduction of the deconvolution layers contributes significantly to reducing computational resource consumption without degrading the accuracy of this system. We also incorporated recent model agnostic techniques such as DarkPose and distillation training to maximize the efficiency of our model. Furthermore, we applied model quantization to exploit multi/mixed precision features. Our FP16'ed model (COCO AP 70.0) operates at ~60-fps on NVIDIA Jetson AGX Xavier and ~200 fps on NVIDIA Quadro RTX6000.

## I. INTRODUCTION

Most visual recognition tasks have advanced significantly owing to the recent evolution of CNNs. Human pose estimation is one such task, which has a wide range of practical applications. Specifically, pose estimation is a vital technique for human activity analysis. It is used not only for 2D and 3D key-point coordinate estimation (i.e., skeleton estimation), but also for action and gesture recognition, as well as person re-identification.

Pose estimation methods are categorized into two groups: bottom-up and top-down. Top-down methods adopt a two-step approach. They first detect person bounding boxes in an image using a person detector and then conduct pose estimation for every detected bounding box. The top-down approach has advantages in terms of accuracy, but it has computational challenges when multiple persons are detected because it requires repeated pose estimation processes. This is a critical issue considering edge and cloud computing, wherein the absolute computational resources are limited or the requirement for additional resources leads to increased costs.

Let us consider passenger monitoring for the 10-fps use-case. Here, a system is required to monitor in-vehicle passengers every 100 ms. If five passengers are on the vehicle, one-person detection and five pose estimations must be completed in a 100 ms window. This is a practical example in which a highly efficient pose estimation system is required. Another use-case is the consolidation of multi-camera image analyses, as found in building and shuttle bus surveillance systems. They must process many human activities in parallel.

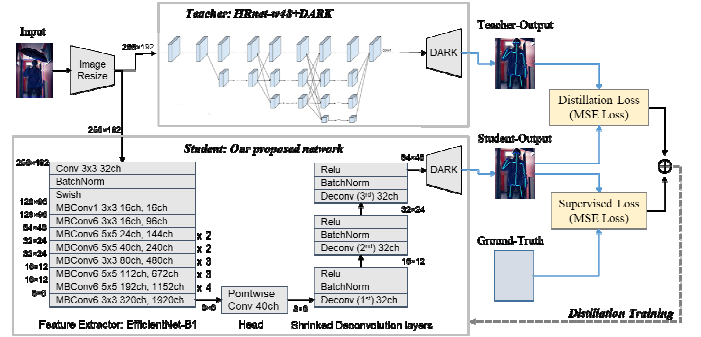

Fig. 1. Our proposed method with shrinking deconvolution achieved via DarkPose and distillation training to achieve an accurate and lightweight.

TABLE I.   2D POSE ESTIMATION METHODS TESTED ON COCO TEST2017 DATASET. OUR MODEL HAS EXHIBITED 0.56 GFLOPS, AP 70.1.

| Method | Input size | GFLOPs | Accuracy | | | |
|---|---|---|---|---|---|---|
| | | | *AP* | *AP50* | *APM* | *AR* |
| OpenPose [2] | - | - | 61.8 | 84.9 | 57.1 | 66.5 |
| HRNet-W48 [3] | 384×288 | 32.9 | 75.5 | 92.5 | 71.9 | 80.5 |
| HRNet-W48+DARK [4] | 384×288 | 32.9 | 76.2 | 92.5 | 72.5 | 81,1 |
| HRNet-W48 [3] | 256×192 | 14.61 | 74.2 | 92.4 | 70.9 | 79.5 |
| Simple Baseline [5] | 256×192 | 8.9 | 70.0 | 90.9 | 66.8 | 75.6 |
| Simple Baseline+DARK [4] | 128×96 | 3.9 | 63.1 | 86.2 | 61.3 | 70.0 |
| IKDnet [6] | 256×128 | 3.2 | 70.3 | - | - | 74.1 |
| EfficientPose-C [7] | 256×192 | 1.6 | 70.9 | 91.3 | 67.7 | 76.5 |
| LPN-101[8] | 256×192 | 1.4 | 70.0 | 90.8 | 66.8 | 75.6 |
| EfficientPose-B [7] | 256×192 | 1.1 | 70.5 | 91.1 | 67.3 | 76.1 |
| LPN -50 [8] | 256×192 | 1.0 | 70.7 | 88.9 | 67.9 | 76.7 |
| **Our model** | **256×192** | **0.56** | **70.1** | **90.9** | **66.9** | **81.2** |

TABLE II.   2D POSE ESTIMATION METHODS TESTED ON COCO VAL2017 DATASET. OUR MODEL HAS EXHIBITED 0.56 GFLOPS, AP 70.5.

| Method | Input size | GFLOPs | Accuracy | | | |
|---|---|---|---|---|---|---|
| | | | *AP* | *AP50* | *APM* | *AR* |
| HRNet-W48 [3] | 256×192 | 14.61 | 75.1 | 90.6 | 71.2 | 80.3 |
| Simple Baseline [5] | 256×192 | 8.9 | 70.4 | 88.6 | 67.1 | 76.3 |
| HRNet-W32+DARK [4] | 128×96 | 1.8 | 70.7 | 88.9 | 67.9 | 76.7 |
| EfficientPose-C [7] | 256×192 | 1.6 | 71.3 | - | - | - |
| LPN -101[8] | 256×192 | 1.4 | 70.4 | 88.6 | 67.2 | 76.2 |
| EfficientPose-B [7] | 256×192 | 1.1 | 71.1 | - | - | - |
| LPN -50 [8] | 256×192 | 1.0 | 69.1 | 88.1 | 65.9 | 74.9 |
| EfficientPose-A [7] | 256×192 | 0.5 | 66.5 | - | - | - |
| **Our model** | **256×192** | **0.56** | **70.5** | **88.8** | **66.8** | **77.3** |

The main purpose of our study is to design a pragmatic pose estimation process that can process 5 to 10 persons in real time using common embedded hardware while maintaining SOTA comparable accuracy.

HRNet [3] is a SOTA method that adopts a top-down approach. It is widely used in many competitions and applications because of its high accuracy and flexibility to accommodate further algorithms such as action recognition. However, HRNet has a multilevel hierarchical structure and complex multi-connections. This method consumes a large amount of computational resources associated with its deconvolution layers. Therefore, it is difficult to make it portable and operate in real-time on low-power hardware.

In this study, we propose a lightweight pose estimation model that we downsize with careful consideration of the deconvolution layers. Our model is portable to deploy on hardware with limited computational resources when operated in real-time while retaining a high accuracy. We adopt recent model-agnostic techniques including DarkPose [4] and distillation training [9]. It can help improve the accuracy of extremely small models such as ours.

## II. RELATED RESEARCH

OpenPose [2] is the most widely used bottom-up-based pose estimation method. It first detects all candidate parts belonging to all people in an entire image and then estimates the probable connection of various parts to construct skeletons. The bottom-up approach operates faster than top-down approach because the bottom-up approach processes an entire image at once, even when multiple people exist within the image. However, estimating the correct connections of multi-person parts is not easy, and it is much worse than that of SOTA top-down approaches. Several downsized variants of OpenPose, such as Lightweight OpenPose [10] and TF-pose [11] have much lower accuracies.

In contrast, the top-down approach first detects person-bounding boxes in an image using a person detector and then conduct pose estimation for every detected bounding box. HRNet and Simple baseline [5] are representative of the top-down approach. The top-down approaches have an advantage over bottom-up approaches in terms of accuracy because it can assume that there is only one person in a bounding box and focus on estimating the key points composing a single person. However, the top-down approach encounters computational challenges when multiple people are detected because pose estimation must be conducted repeatedly on all bounding boxes. HRNet and Simple Baseline adopt the encoder–decoder model, which is composed of down-sampling followed by up-sampling processes. This structure is known as an effective architecture for pose estimation. However, the up-sampling layers require tensor scaling or deconvolution, which usually consumes large amounts of computing resources and becomes a major performance bottleneck. In addition, several high-performing top-down models use uncommon layers, computationally demanding additional mechanism and hierarchical skip connections, which make them unsuitable to embed on cheap and low-power consumption hardware.

BlazePose [12] is a lightweight top-down approach that realizes real-time pose estimation using a mobile phone CPU. Although it is lightweight, its accuracy is comparable to OpenPose. EfficientPose [7] incorporates EfficientNet [13], which is based on the feature extractors generated by network architecture search (NAS). Zhang et al. [4] indicated that key-point mapping from the model output to a final target resolution has a significant impact on the accuracy of these methods. They proposed a principal method to reflect the model output distribution (heat map) to the final resolution using Taylor expansion, which they referred to as DarkPose. In general the most common method to reduce the computational burden of the top-down approach such that they may be applied towards embedded applications is to reduce the input image size in exchange for significant accuracy degradation.

## III. PROPOSED METHOD

Our proposed method is composed of a simple encoder–decoder architecture, which results in an architecture that is suitable for embed in common edge computers such as NVIDIA Jetson AGX Xavier and Xilinx FPGA. We ensured between-hardware portability using only common layers and parameters. We adopted a top-down approach based on Simple Baseline and incorporated EfficientNet, DarkPose, and applied distillation training to design the lightweight model while minimizing the accuracy degradation.

To date, the optimal design of deconvolution layers has not yet been thoroughly studied. We investigated the deconvolution layer composition in great detail to identify the most efficient design and found that the channel reduction of the deconvolution layers contributes significantly to reducing resource consumption without degrading the accuracy.

### A. Designed Model

As shown in Fig. 1, we adopted an encoder–decoder model similar to that of Simple Baseline as our base model and fine-tuned this model such that it may be utilized with embedded hardware with scarce resources. It is common to use a downsized resolution as an input to these models to reduce the required amount of computation. Pose estimation models typically use resolutions such as H256×W192, and the feature extractor down-samples these resolutions to H8×W6. Next, the input is up-sampled (i.e., to H64×W48) using deconvolution layers. The model output must then be mapped to the original resolution, and we used DarkPose to calculate the output of our model. For training, we used HRNet+DarkPose as the teacher model for distillation training. The mean squared error is used to determine the distillation loss and supervised loss. We designed a number of channels to possess multiples of eight in all layers to ensure hardware computation efficiency.

### B. Feature Extractor

We adopted EfficientNet which originally has the optimal combination of network width, depth, and input resolution. However, we needed to adjust the combination to fit the model on resource limited embedded hardware.

## C. Shrinking Deconvolution Layers

The number of channels in the deconvolution layers greatly affects the computational costs. We attempted to utilize model configurations with varying numbers of channels. The layer number of the deconvolution layers is another important factor that affects both computational cost and accuracy. Xiao et al. [5] identified that the use of three layers produced good results, and we restudied the model configuration with different layer numbers to embed it into the selected hardware.

## IV. EXPERIMENTS

As shown in Tables I and II, our method achieved very good accuracy with an extremely low calculation cost. Our best model achieved an AP of 70.5 on COCO val2017 dataset [1] with 0.56 GFLOPs. It is also quantization-ready, and the FP16'ed model is implementable on NVIDIA Jetson AGX Xavier. The accuracy degradation of our model from SOTA HRNet-W48 256×192 (AP 74.2%) on the COCO test2017 dataset was suppressed to 4.1 pt (5.5%) while the number of calculations was reduced to 3.8%. In comparison, OpenPose has an AP of only 61.8. In the top-down approach, 8-stage Hourglass [14] and HRNet achieved higher accuracies, but these methods are hindered by their high computational complexity. EfficientPose-A is nearly the size of our model, but its accuracy is significantly worse. In the following sections, we report upon the fine-tuning of our optimal results.

TABLE III. BENCHMARK INPUT SIZE ON COCO VAL2017 DATASET

| Input size | GFLOPs | Accuracy (AP) |
|---|---|---|
| 128×96 | 0.14 | 54.8 |
| 160×128 | 0.24 | 61.4 |
| 192×160 | 0.35 | 65.2 |
| 224×160 | 0.41 | 67.3 |
| **256×192** | **0.56** | **69.7** |
| 384×288 | 1.27 | 72.9 |

TABLE IV. BENCHMARK ARCHITECTURE ON COCO VAL2017 DATASET

| Feature Extractor Input Size 256×192 | Deconvolution [channel] 1st | 2nd | 3rd | GFLOPs | Accuracy (AP) |
|---|---|---|---|---|---|
| Resnet-18 | 16 | 16 | 16 | 1.69 | 48.3 |
| Resnet-18 | 32 | 32 | 32 | 1.76 | 66.9 |
| Resnet-18 | 64 | 64 | 64 | 1.99 | 67.1 |
| Resnet-18 | 128 | 128 | 128 | 2.79 | 67.7 |
| Resnet-18 | 256 | 256 | 256 | 5.79 | 67.4 |
| Resnet-34 | 32 | 32 | 32 | 3.45 | 70.7 |
| Resnet-50 | 32 | 32 | 32 | 3.98 | 71.2 |
| Resnet-50 | 256 | 256 | 256 | 8.99 | 71.6 |
| EfficientNet-B0 | 32 | 32 | 32 | 0.40 | 67.2 |
| **EfficientNet-B1** | **32** | **32** | **32** | **0.56** | **69.7** |
| EfficientNet-B2 | 32 | 32 | 32 | 0.64 | 69.8 |
| EfficientNet-B3 | 32 | 32 | 32 | 0.92 | 7.11 |
| EfficientNet-B4 | 32 | 32 | 32 | 1.40 | 7.13 |
| EfficientNet-B5 | 32 | 32 | 32 | 2.17 | 7.18 |
| EfficientNet-B6 | 32 | 32 | 32 | 3.07 | 72.5 |

## A. Tuning Parameters

We trained our model using COCO train2017 dataset with data augmentation involving random image rotation and scaling. The batch size was 64 for each GPU. We used the Adam optimizer and trained it for a total of 140 epochs. We used 1e-3 as the initial learning rate, which was then decreased to 1e-4 for 90 to 139 epochs, and finally to 1e-5 for 120 to 140 epochs. Our model has 12.26 M weight parameters.

## B. Input Resolution

Table III lists the accuracy and computational cost variations for various input resolutions. Input resolution is a typical trade-off parameter between the accuracy and computational cost of these systems. The input resolution should be selected based on the specific application requirements. We selected the H256×W192 resolution for our representative model.

## C. Feature Extractor

Table IV shows the evaluation process using different sizes of EfficientNet and ResNet [15]. We confirmed the advantages of using EfficientNet as a feature extractor compared to ResNet. Although the selection of the model size among the EfficientNet family represents another trade-off between the accuracy and computational cost, we found that the accuracy benefit is not large and smaller sizes may be acceptable depending on the application. We selected EfficientNet-B1.

## D. Deconvolution Layers

As shown in Table V, we found that the number of channel reduction of the deconvolution layers contributes significantly to reducing the amount of required computation, while it does not appear to linearly degrade the accuracy. As shown in Table VI, adding more than three layers did not improve the system accuracy. We selected a three-level hierarchical architecture composed of three deconvolution layers and 40 channels for an extractor head, as well as 32-32-32 channels for deconvolution layers because of the accuracy of this architecture. U-Net [16] shows that a attention connection improves the accuracy of the encoder–decoder model, but we found that the skip connection does not significantly improve the accuracy of the pose estimation, as shown in Table VII.

## E. Distillation Training

Distillation learning is also effective for pose estimation. We trained our model (EfficientNet-B1, three-level hierarchical architecture, 40-channel head, and 32-32-32 channels for the deconvolution layers) using HRNet-W48-DarkPose (AP 75.0, 15.77 GFLOPs, 256×192) as the teacher for distillation training. The distillation training improved the accuracy of AP +0.8 pt from AP 69.7 to AP 70.5 on COCO val2017 dataset.

## V. PERFORMANCE ON HARDWARE

Model portability among various hardware platforms is crucial for practical applications. We carefully designed our model to use only selected common layers so that it could be integrated into the target hardware without any changes or with small modifications. We assumed the use of NVIDIA Jetson AGX Xavier and Xilinx FPGA as the common hardware and embedded our model into these systems.

## A. Embedding into NVIDIA Jetson AGX Xavier

We quantized the model precision from FP32 to FP16/INT8 with NVIDIA TensorRT (latest version 7.2) optimization tool to embed our model and studied its effect on the accuracy of our model. Table VIII shows the accuracy and inference time of the optimization results FP32 precision of our baseline model (AP 70.5, 115.56 ms) was quantized to FP16 precision (AP 70.0, 15.57 ms) and INT8 precision (AP 66.2, 22.57 ms). INT8 precision has a relatively large accuracy loss.

Regarding the inference time, the FP16'ed model is faster benefit from Tensor Core acceleration, which is faster than CUDA Core. However, FP32 and INT8'ed models could not leverage Tensor Core according to the Tensor RT optimization outputs, and these inference times are therefore longer. Regarding the accuracy, INT8 quantization decreases AP -4.3 pt from FP32 precision, which is larger than the decrease achieved via FP16 quantization. For applications that depend on computation time, the FP16'ed model is a practical solution.

## B. Embedding into Xilinx FPGA

As shown in Fig. 2, we used the ReLU activation layer instead of the Swish activation layer and removed the SE-block to achieve hardware portability. While the original EfficientNet adopts the optimized combination of input resolution, layer number, width, and depth, we further reduced each parameter of the layers to fit the more restrictive Xilinx FPGA specifications compared to those of NVIDIA Jetson AGX Xavier. As shown in Fig. 3, we defined the specifications of the model to achieve efficient computing by referring to Xilinx FPGA (DPU B1024). The developed model exhibited AP 66.7 and 0.46 GOPs on COCO val2017 dataset. A degraded accuracy was observed as the number of channels was reduced. The accuracy was 14.12 ms for the FP16'ed model on Xilinx DPU B2304 in the Avnet Ultra96-V2 FPGA board and 19.83 ms on B1024, which has half the computing power of B2304. The Ultra96-V2 system can possess two B1024s and process 5 to 10 people in real time.

## VI. CONCLUSION

We present a high accuracy lightweight pose estimation model. We assumed that the use of Taylor interpolation during post-processing enabled our model to utilize a reduced number of channels without severe accuracy degradation. We also applied distillation training, which are less dependent on the model structure, to improve the overall accuracy of an estimation model. As a result, we reduced the weight of the developed model to 0.56 GFLOPs while achieving nearly the same accuracy as other methods that are more than twice the size (1.0 GFLOPs) of the proposed method. Our report will benefit the further efficient optimization of encoder-decoder methods and multi-level structures such as HRNet, which is often used for pose estimation and action/gesture recognition.

TABLE V. BENCHMARK NUMBER OF DECONVOLUTION LAYERS ON COCO VAL2017 DATASET

| Multi-level Hierarchy | GFLOPs | Accuracy (AP) |
|---|---|---|
| 2 | 0.37 | 67.1 |
| **3** | **0.56** | **69.7** |
| 4 | 0.75 | 68.6 |

TABLE VI. BENCHMARK CHANNEL PARAMETERS OF DECONVOLUTION USING OUR REDUCED EFFICIENTNET-B1 ON COCO VAL2017 DATASET

| Head [channel] | Deconvolution [channel] | | | GFLOPs | Accuracy (AP) |
|---|---|---|---|---|---|
| | *1st* | *2nd* | *3rd* | | |
| 320 | 256 | 256 | 256 | 4.50 | 69.6 |
| 160 | 128 | 128 | 128 | 1.50 | 69.0 |
| 80 | 64 | 64 | 64 | 0.75 | 69.2 |
| **40** | **32** | **32** | **32** | **0.56** | **69.7** |
| 20 | 16 | 16 | 16 | 0.52 | 64.8 |
| 10 | 8 | 8 | 8 | 0.50 | 19.7 |

TABLE VII. BENCHMARK MULTI-LEVEL HIERARCHICAL CONVOLUTION AND DECONVOLUTION LAYERS ON COCO VAL2017 DATASET

| Attention Type | GFLOPs | Accuracy (AP) |
|---|---|---|
| **Non connection** | **0.56** | **69.7** |
| Concat | 0.56 | 69.4 |
| Sum | 0.56 | 68.4 |

TABLE VIII. PRECISION OF THE MODEL ACCURACY ON COCO VAL2017 DATASET, AS WELL AS THE INFERENCE TIME WITH DIFFERENT HARDWARE

| Precision | Accuracy (AP) | Inference Time | |
|---|---|---|---|
| | | NVIDIA Jetson Xavier AGX | NVIDIA Quadro RTX6000 |
| FP32 | 70.5 | 115.56 | 19.10 |
| FP16 | 70.0 | 15.57 | 4.96 |
| INT8 | 66.2 | 22.57 | 4.17 |

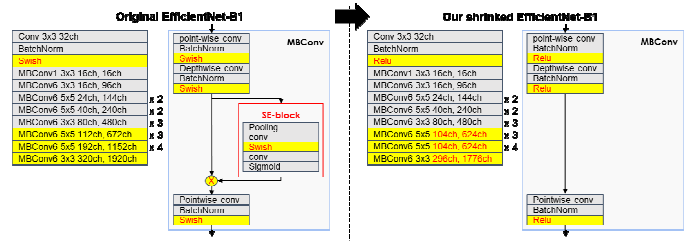

Fig. 2. Shrinking EfficientNet-B1 as a feature extraction method with common layers and light-weight parameters (see Fig. 3) for embedding within common hardware such as NVIDIA Jetson AGX Xavier and Xilinx FPGA.

Fig. 3. Defined specifications of our model to achieve efficient computing on common hardware, as referenced from Xilinx FPGA (DPU B1024), which is supported by MobileNet [17] with a low-power consumption CNN accelerator.

## REFERENCES

[1] T. Lin, M. Maire, S. Belongie, J. Hays, P. Perona, D. Ramanan, P. Dollar, and C. Lawrence, "Microsoft COCO: Common objects in context" in ECCV, 2014, pp. 740–755.

[2] Z. Cao, T. Simon, S. Wei, and Y. Sheikh, "Realtime Multi-Person 2D Pose Estimation using Part Affinity Fields," in CVPR, 2017, pp.1302–1310.

[3] K. Sun, B. Xiao, D. Liu, and J. Wang, "Deep High-Resolution Representation Learning for Human Pose Estimation." in CVPR, 2019.

[4] F. Zhang, X. Zhu, H. Dai, M. Ye, C. Zhu, "Distribution-Aware Coordinate Representation for Human Pose Estimation," in CVPR, 2020.

[5] B. Xiao, H. Wu, and Y. Wei, "Simple baselines for human pose estimation and tracking," in ECCV, 2018, pp. 472–487.

[6] X. Xu, Q. Zou, X Lin, Y Huang, and Y Tian, "Integral Knowledge Distillation for Multi-Person Pose Estimation," IEEE Signal Processing Letters 27, 2020, pp.436-440.

[7] W. Zhang, J. Fang, X. Wang, W. Liu, "EfficientPose: Efficient Human Pose Estimation with Neural Architecture Search," in Computational Visual Media, 2021.

[8] Zhe Zhang, Jie Tang, Gangshan Wu, "Simple and lightweight human pose estimation," in arXiv:1911.10346, 2019.

[9] G. E. Hinton, O.Vinyals, and J. Dean, "Distilling the knowledge in a neural network," in Proc.NIPS Deep Learn. Workshop, 2014.

[10] Daniil Osokin, "Real-time 2D Multi-Person Pose Estimation on CPU: Lightweight OpenPose," in arXiv:1811.12004, 2018.

[11] Z. Cao, "tf-pose-estimation," https://github.com/ZheC/tf-pose-estimation, unpublished.

[12] V. Bazarevsky, I. Grishchenko, K. Raveendran, T. Zhu, F. Zhang, M. Grundmann, "BlazePose: On-device Real-time Body Pose tracking," in CVPR, 2020.

[13] M. Tan and Q. V. Le, "Efficientnet: Rethinking model scaling for convolutional neural networks," in The International Conference on Machine Learning, 2019.

[14] A. Newell, K. Yang, and J. Deng, "Stacked hourglass networks for human pose estimation," in ECCV, 2016, pp. 483–499.

[15] He, K., Zhang, X., Ren, S., and Sun, J., "Deep residual learning for image recognition," in CVPR, 2016, pp. 770–778.

[16] Olaf Ronneberger, Philipp Fischer, and Thomas Brox, "UNet: Convolutional networks for biomedical image segmentation," in Proc. Medical Image Computing and Computer Assisted Intervention, pp. 234–241, 2015.

[17] Andrew G. Howard, Menglong Zhu, Bo Chen, Dmitry Kalenichenko, Weijun Wang, Tobias Weyand, Marco Andreetto, and Hartwig Adam, "Mobilenets: Efficient convolutional neural networks for mobile vision applications," in arXiv:1704.04861, 2017.